\documentclass[12pt]{article}
\usepackage[defblank]{paralist}
\usepackage[tbtags]{amsmath}
\usepackage{amssymb,theorem}
\usepackage{nicefrac}
\usepackage{titlesec}
\usepackage{multicol}
\usepackage[dvips]{graphicx}
\usepackage{mathrsfs}
\usepackage{url}
\usepackage[a4paper]{geometry}
\usepackage{color}
\usepackage{psfrag}
\usepackage{array}
\usepackage{booktabs}
\usepackage{epic}
\usepackage{eepic}
\usepackage{subfigure}
\usepackage{verbatim} 
\usepackage{colortbl}
\usepackage{tikz}
\usepackage{pgf}

\usepackage[authoryear]{natbib}

\usetikzlibrary{automata,arrows,shapes,snakes,topaths,trees}

\DeclareGraphicsExtensions{.jpg,.pdf,.png}
\newcommand{\beq}{\begin{equation}}
\newcommand{\eeq}{\end{equation}}
\newcommand{\bdm}{\begin{displaymath}}
\newcommand{\edm}{\end{displaymath}}
\newcommand{\bea}{\begin{eqnarray*}}
\newcommand{\eea}{\end{eqnarray*}}
\newcommand{\bdes}{\begin{description}}
\newcommand{\edes}{\end{description}}
\newcommand{\bit}{\begin{itemize}}
\newcommand{\eit}{\end{itemize}}
\newcommand{\benum}{\begin{enumerate}}
\newcommand{\eenum}{\end{enumerate}}
\newcommand{\bc}{\begin{center}}
\newcommand{\ec}{\end{center}}

\newcommand{\mc}{\mathcal}

\newcommand{\qed}{\nobreak \ifvmode \relax \else
\ifdim\lastskip<1.5em \hskip-\lastskip
\hskip1.5em plus0em minus0.5em \fi \nobreak
\vrule height0.75em width0.5em depth0.25em\fi}

\title{How a minimal learning agent can infer the existence of unobserved variables in a complex environment}
\begin{document}
\date{\today}
\author{Katja Ried, Benjamin Eva, Thomas M\"uller, Hans J. Briegel}
\maketitle
\abstract{According to a mainstream position in contemporary cognitive science and philosophy, the use of abstract compositional concepts is both a necessary and a sufficient condition for the presence of genuine thought. In  this  article,  we  show  how  the  ability to develop and utilise abstract conceptual structures  can  be  achieved  by  a particular kind of 
learning  agents. More  specifically,  we  provide and motivate a concrete operational definition of what it means for these agents to be in possession of abstract concepts,  before presenting an  explicit  example of  a  minimal  architecture  that  supports  this  capability.   We  then  proceed  to  demonstrate  how the existence of abstract conceptual structures can be operationally useful in the process of employing previously acquired knowledge in the face of new experiences, thereby vindicating the natural conjecture that the cognitive functions of abstraction and generalisation are closely related.

}

\section{Introduction}

\subsection{Objectives}

According to a mainstream position in contemporary cognitive science and philosophy, the use of abstract compositional concepts is both a necessary and a sufficient condition for the presence of genuine thought (see, e.g., \cite{Bermudez2003Thinking}, \cite{Carruthers2009Invertebrate}, \cite{Evans1982Varieties}). Indeed, the verifiable possession of compositional concepts is widely forwarded as a criterion that needs to be satisfied before any substantive doxastic states can be legitimately attributed to non-human animals (see, e.g., \cite{Carruthers2009Invertebrate, Davidson1975Thought, Dreyse2011Honeybees}). If one takes this kind of position seriously (as many do), it follows that any system genuinely deserving of the name `artificial intelligence' will possess the ability to effectively traffic in abstract conceptual representations of salient features of its environment. (Indeed, numerous variations of this view have already been articulated and defended in the foundations of AI literature, e.g., \cite{Bengio2013Representation, Lake2015Human}). In this paper, we address this observation by constructing an explicit example of a simple learning agent that autonomously identifies abstract variables in the process of learning about its environment, before providing a concrete operational semantics that allows external observers to subsequently identify these variables through analysis of the agent's internal deliberative structures. Moreover, we demonstrate how an agent's ability to construct and employ these abstract conceptual structures correlates with its ability to employ previously acquired knowledge when dealing with novel experiences. 

Beyond the motivation of constructing AI systems that satisfy the criterion of possessing abstract compositional conceptual structures, we take the significance of this work to be twofold. Firstly, by constructing learning agents that are capable of discovering abstract variables in a way that can be objectively identified in subsequent analysis, we take a meaningful step towards developing artificial agents whose reasoning processes are fully transparent, interpretable and communicable. Unlike conventional reinforcement learning algorithms, which do not develop any discernible conceptual structures and do not support any meaningful interpretation of what they have learned (see, e.g., \cite{SuttonBarto1998, WieringOtterlo2012}), the agent proposed in the present work structures the information that it gathers in a way that supports an operational interpretative semantics, which is an important first step towards combining the efficient learning abilities of reinforcement learning agents with explicit and communicable symbolic deliberations. 

The second point of significance is our observation that agents that have identified abstract variables perform noticeably better at tasks that require them to generalise existing knowledge to deal with new experiences. This both provides a novel operational vindication for the pragmatic and epistemic value of abstract conceptual representations, and solves an existing operational problem regarding the ability of reinforcement learning agents to successfully generalise. 

The paper is structured as follows. In section \ref{sec:learning_env}, we describe the kind of learning environment used to investigate the formation and identification of abstract variables. In section \ref{sec:learning_agent}, we introduce the particular type of reinforcement learning agent to be deployed in those tasks (namely `Projective Simulation' agents) before presenting 
a novel modification to the architecture of those agents, which provides them with the necessary `cognitive space' for variable identification. In section \ref{sec:var_ident} we formally specify what it means for 
 such an agent to identify variables in the context of the learning tasks described in section \ref{sec:learning_env}. In section \ref{sec:results_main} we present the results of our simulations, which illustrate the efficacy of our variable identification protocol. In particular, subsection \ref{sub:results_gen} analyses the observed correlation between the existence of identifiable variables in an agent's deliberations and the ability of that agent to deal with novel experiences in an effective manner. Section \ref{sec:discussion} concludes.  

\section{The Learning Environment}
\label{sec:learning_env}

\subsection{Basic Structure}
\label{subsec:env_basic}

Our central aims are (i) to enable a learning agent to infer the existence of unobserved variables in a complex environment via dynamic interactions, and (ii) to subsequently develop an operational semantics that allow us to identify a representation of these variables in the agent's internal deliberation structures. Towards this end, we consider an environment that consists initially of three components:

\begin{itemize}
\item A set $\mc{S}$ of possible \emph{setups}, i.e., situations on which experiments can be performed. For example, in a context in which the agent is allowed to perform simple classical physical experiments on a range of different objects, each setup $s \in \mc{S}$ could represent one object. More generally, each setup $s$ represents a different situation that the agent can test via a range of experiments, such that each situation can be distinguished by the results it yields in at least some of the available experiments.
\item A set $\mc{E}$ of \emph{experiments}, i.e., tests that can be performed on any of the available setups. For example, in the case in which the agent can perform classical physical experiments on objects, one possible experiment could be 
suspending a given object from a spring and recording by how much the spring is extended.
\item A set $\mc{P}$ of \emph{predictions} such that each $p \in \mc{P}$ corresponds to a prediction of the outcome of exactly one experiment in $\mc{E}$. For example, if one of the available experiments is to measure the spin of a particle along the $y$-axis, then $\mc{P}$ would contain one prediction corresponding to the `spin down along the $y$-xis' outcome and one prediction corresponding to the `spin up along the $y$-axis' outcome. 
\end{itemize}

A few additional comments regarding the predictions are in order. Since, by assumption, each prediction $p\in\mc{P}$ corresponds to exactly one experiment $e\in\mc{E}$, one can think of  $p$ as including a specification of which experiment it pertains to. A prediction $p$ is then deemed `correct' for a given setup $s$ if, under the experiment $e$ for which $p$ is a possible prediction, the setup $s$ indeed produces the corresponding outcome. Note that, in what follows, we make the simplifying assumption (to be relaxed in future work) that the outcomes of experiments are deterministic, i.e., that each setup/experiment pair predetermines a unique correct prediction.
Moreover, we assume that $\mc{P}$ contains a complete set of the possible outcomes for every $e \in \mc{E}$, in the sense that there can be no combination of a setup $s \in \mc{S}$ and an experiment $e \in \mc{E}$ performed on it such that the resulting outcome is not among the predictions $\mc{P}$. (One can ensure that this holds true even in pathological cases, such as an attempt to measure the spin of a particle in the eventuality that no particle is present, by formally including one prediction to the effect of `not applicable'.) 

Once the environment has been fully specified via a choice of $\mc{S}, \mc{E}$ and $\mc{P}$, agents interact with it in the following way. Each round of interaction begins with the agent being presented with a single setup $s \in \mc{S}$ that is drawn from a fixed probability distribution over $\mc{S}$, which we assume to be uniform.\footnote{Again, the uniformity of the distriution over setups is a simplifying assumption that will be relaxed in future work. We restrict ourselves to the simplest case here in order to focus on the central theme of conceptual abstraction without unnecessary technical distractions.} Upon being presented with $s$, the agent is asked to make a prediction $p \in \mc{P}$ (which, as detailed above, implicitly includes a choice of an experiment $e$). 
Finally, the agent receives a reward if and only if their prediction is correct for the setup. 
For example, the agent could be presented with a particular object $s$ and asked to make a prediction for any one of the available experiments that could be performed on that setup, i.e., placing it on a scale. They would then be rewarded if and only if their prediction matched the outcome of that experiment, i.e., the readout on the scale.

The above learning environment is reminiscent of classic reinforcement learning tasks, in which success is equated with efficiently learning how to choose the correct option (prediction) for all possible inputs (setups), i.e., with efficiently learning how to maximise rewards. However, in our approach, this is only the first step of what constitutes successful learning. Rather than merely learning how to make correct predictions, our central success criterion is that the agent develop transparent and easily interpretable conceptual representations of those aspects of their environment that play a role in determining the outcomes of experiments.

To make this success criterion precise, we will introduce one additional component in our description of the environment. It is based on the observation that each setup could be uniquely identified by a specification of the values of a number of suitable abstract variables, e.g., the size, shape and composition of an object. Crucially, we do not assume that the agent can perceive the values of these variables, or even that they are aware of the fact that a description of the observed setup can be compressed in such a way. Our goal is precisely to construct an agent that can infer the existence of such variables even if the setups are presented as mere atomic labels that carry no intrinsic meaning. Formally, we assume that, in addition to the three components $\mc{S, E, P}$ specified above, the environment also contains a `hidden' fourth component, namely

\begin{itemize}
    \item A set $\mc{V}$ of `hidden' (or latent) variables, i.e., variables which are never explicitly presented to the agent, but which are sufficient to determine the outcomes of all experiments.\footnote{We stress that that `hidden variable' terminology is not intended to reflect the usage of the terminology in the quantum foundations community. By `hidden variable', we mean simply an environmental variable that is not directly observable for the agent.} Each setup $s \in S$ can be equated with a vector specifying exactly one value for each of the variables in $\mc{V}$, and each experiment is assumed to test one and only one of the variables in $\mc{V}$, although there may be multiple experiments testing the same variable. For instance, if there are two variables with two values each, then there will be four classes of setups corresponding to the four possible configurations of the values of the variables in $\mc{V}$, i.e $s_{1} = 00, s_{2} = 01, s_{3} = 10, s_{4} = 11$. There will also be at least two experiments, each corresponding to one variable, where the outcome of each experiment is determined by the value that the given setup entails for the corresponding variable. 
\end{itemize}

The problem of unobserved variables is also relevant to the field of machine learning (specifically reinforcement learning), in the context of partially observable Markov decision processes (POMDPs, see, e.g., \cite{Poupart2012POMDP}). 
In such processes, the input available to the agent does not contain sufficient information to completely characterise the state of the environment, or to make deterministic assessments of the consequences of possible actions. By contrast, in the scenario considered here, the input (the setup $s$) does completely specify the state, in the sense that $s$ determines with certainty the outcomes of all possible experiments that could be performed on it. What is unavailable to the agent are merely auxiliary variables that help \emph{structure} the relations between the various setups in $\mc{S}$ and the corresponding predictions.

\subsection{Concrete scenario used in training agents}
\label{sub:standard_scenario}

\begin{figure}
\begin{centering}
\includegraphics[width=.99\textwidth]{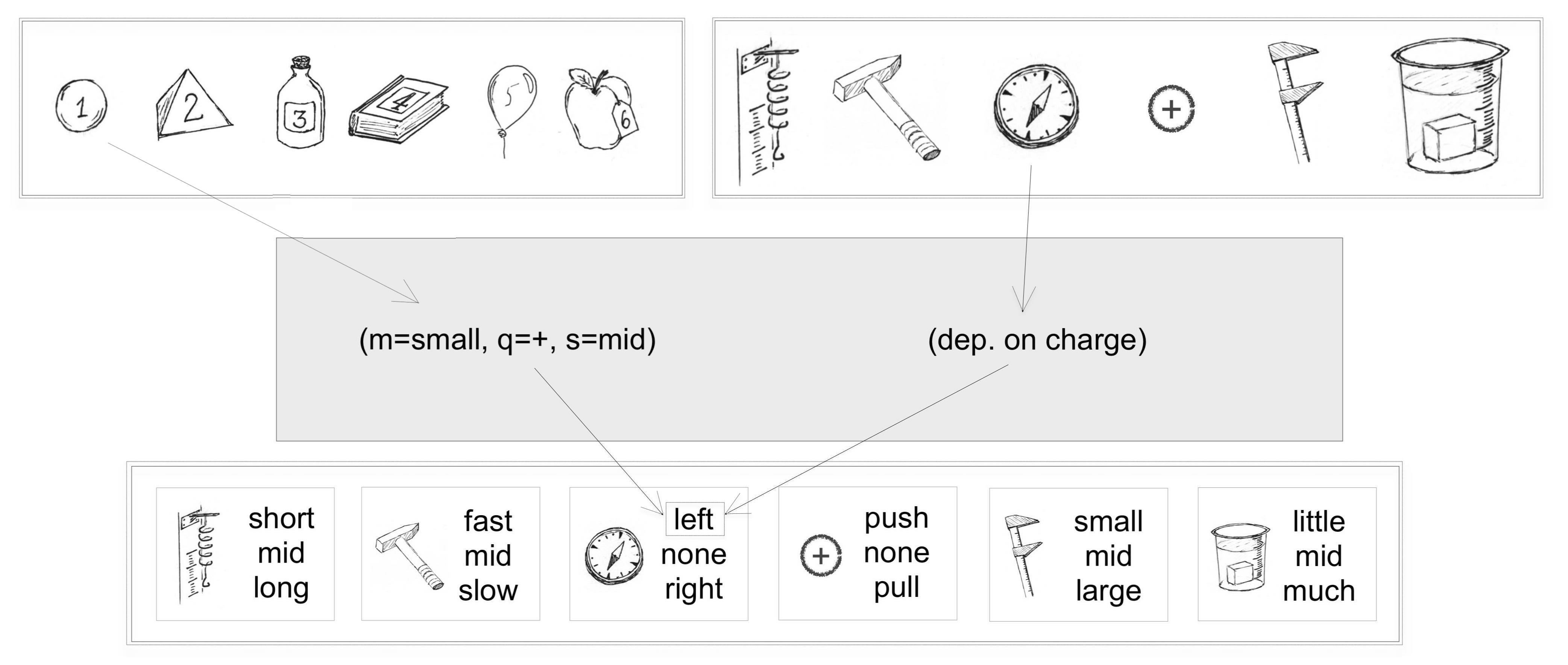}
\par\end{centering}
\caption{\label{fig:environment}
A task environment with hidden variables and rules to be discovered: the agent receives objects (top left) on which it can perform a range of experiments (top right), whose outcomes (bottom) it attempts to predict. What is hidden from the agent (grey box) is that each object can be described by a vector of values, namely its mass, charge and size, and each experiment can be predicted given the value of one of these variables: when suspending the object from a spring (`scale experiment') or hitting it to impart a given momentum (`momentum experiment'), the outcome depends on the object's mass; when passing it near a compass or placing it near a test charge, the results depend on the object's charge, while the outcomes of grasping the object or submerging it in a bucket of water depend on its size.}
\end{figure}

To illustrate these ideas, we now provide a concrete example of a learning environment containing hidden variables. This scenario will also be used as the default case in our subsequent analysis of the agents' learning capabilities. It is illustrated visually in Fig.~\ref{fig:environment}.

In our default scenario, each setup consists of an object that the agent can experiment on, which is characterised by $|\mc{V}| = 3$ hidden variables: mass, size and electric charge. (Since objects can in principle have different densities, mass and size are generally not correlated.) Each of these properties predicts the outcomes of 2 different experiments: for example, electric charge predicts what force the object will experience when placed next to a known test charge, and also how much the object will deflect a compass needle when moving past it at a given speed and distance. Mass predicts what will happen if an object is hit, imparting a fixed amount of momentum, or if it is placed on a scale in a known gravitational field, and similarly an object's size predicts its behaviour in two other experiments. Overall, this environment admits $|\mc{E}| = 6$ experiments that the agent can perform. We assume that the possible values of each variable are coarse-grained into 3 distinct values; for example, the variable `electric charge' can take the (coarse-grained) values `positive', `negative' and `neutral' and the variable `size' can take the values `big', `small' and `medium'. These values are reflected in corresponding outcomes for each experiment, so that the number of predictions corresponding to each experiment is also\footnote{Here we assume that whenever an experiment $e$ tests a hidden variable $V$, each possible outcome of $e$ (prediction for $e$) corresponds to exactly one possible value of $V$. It is possible to relax this assumption, but it plays a useful simplifying role in what follows.} 3. This gives rise to a total of $|\mc{P}| = 6\cdot3 = 18$ predictions that the agent can choose from, of which 6 will be correct for any given object. The expected success rate for random guesses in this environment is therefore $1/3$. Since setups (objects) are identified with configurations of the values of the variables in $\mc{V}$, it follows that $|\mc{S}| = 27$, i.e., there are 27 distinct objects on which the agent is able to experiment, and each of those objects instantiates one of the 27 possible configurations of the values of the `charge', `size' and `mass' variables. 

The agent's interaction with this environment proceeds as outlined above: the agent is presented with a randomly chosen object, which is labelled simply with an integer between 1 and 27; the agent then chooses an experiment, makes a corresponding prediction and finally receives a reward if that prediction was correct for the given object. This process is iterated long enough for the agent to eventually encounter all the available objects and learn about them. Specifically, the agents whose results are analysed in the following are given $T=5*10^6$ rounds of interaction with the environment in order to learn, in the default case.

In the standard reinforcement learning paradigm, the criterion for success in the kind of learning task described here would be that the agent successfully learns how to make the correct predictions for all object/experiment pairs. This is a purely operational criterion that can be straightforwardly accomplished in a reinforcement learning setting by implementing a learning dynamics that increases an agent's disposition to make particular predictions in proportion to the extent to which those predictions have been rewarded in the past (and implementing some form of greed avoidance). However, we have also introduced a second criterion for success, which is the central aim of the present work: that the agent, beyond learning how to reliably predict the outcomes for all object/experiment pairs, also comes to identify that there are three hidden variables that determine which predictions will be correct for each object/experiment pair. 

\subsection{The Value of Understanding}

With this formal description of the environment in hand, it is worth pausing to reiterate a few of the central motivations behind this second success criterion. Firstly, identifying the variables at play in the agent's deliberation is a crucial first step towards rendering the agent's deliberations genuinely \emph{transparent}, \emph{interpretable} and \emph{communicable}. Secondly, there is a significant difference between an agent that merely memorises which predictions were rewarded for which object/experiment pairs and an agent that has identified that there exist significant unobserved variables  \textendash{}   which we might identify as `mass', `size' and `charge' \textendash{}  and makes predictions on the basis of which value a given variable takes for a given object. (For example, the agent predicts that the second object will present a reading of `high' in the scale experiment because it 
already knows that (i) there exists a variable that predicts the outcome of the scale experiment (`mass') and (ii) that, based on the momentum experiment, the object has high mass.) 
It is natural to say that the second agent possesses a genuine \emph{understanding} of its environment, in a sense which is absent for the first agent.\footnote{This is closely related to the point that \cite{Block1981Psychologism} makes in his famous `Blockhead' thought experiment, which is intended to refute the Behaviourist conception of intelligence enshrined in the Turing test.} A similar sentiment is expressed by \cite{Bengio2013Representation}, who writes

 \begin{quote}
 An AI must fundamentally understand the world around us, and we argue that this can only be achieved if it can learn to identify and disentangle the
underlying explanatory factors hidden in the observed milieu of low-level sensory data (\cite{Bengio2013Representation}: 1798)
 \end{quote}

Thirdly, many cognitive abilities are grounded in the ability to describe one's environment in terms of abstract conceptual representations. Saliently, the ability to generalise previously acquired knowledge to deal with new experiences seems to be intimately connected to the ability to represent significant properties of one's environment in terms of abstract variables. (This intuitive conjecture is empirically vindicated in section \ref{sub:results_gen}.) More generally, there are numerous pragmatic and theoretical motivations for regarding the identification of abstract variables corresponding to the environment's hidden variables as a success criterion for explorative learning agents. In section \ref{sec:var_ident}, we will provide a concrete formalisation of this second success criterion for a particular kind of reinforcement learning agent. First, we turn to specifying the precise cognitive architecture of those agents.

\section{The learning agent}
\label{sec:learning_agent}

We will work within the context of the projective simulation (PS) framework for artificial intelligence agents, which was first proposed by \cite{BriegelDelasCuevas2012}.
This framework aims to provide a concrete example of what it means to be a deliberating agent: entities that can \emph{act} on their environment, thereby generally changing its state, and, more importantly, that make their \emph{own decisions} in the sense that they are not pre-programmed to take particular actions under given circumstances, but instead are flexible and develop their own action and response patterns. 

While one of the achievements of the PS framework is to provide a concrete, explicit model of agency, the agents' ability to learn has been a point of considerable interest, having been tested against more utilitarian reinforcement learning algorithms on a number of benchmark problems (see, e.g., \cite{BriegelDelasCuevas2012, Mautner2015Comprehensive, Melnikov2018Benchmarking}). The broad conceptual-mathematical basis also supports much more diverse applications, ranging from the autonomous development of complex skills in robotics \citep{Hangl2016Robotic,Hangl2017Skill} through modelling of collective motion in animal swarms \citep{Ried2019Locust} to the control of quantum systems~\citep{Tiersch2015Adaptive, Wallnoefer2019LongDist, Nautrup2018Optimizing} and the design of new experiments~\citep{Melnikov2018Active}.

The interaction of the agent with its environment is formalised following the general framework of reinforcement learning (see, e.g., \cite{SuttonBarto1998}): the learner receives a \emph{percept} that encodes some information about the state of its environment, based on which it chooses an \emph{action}, and, if the action puts the environment in a state that satisfies some pre-defined success criterion, the learner is given a \emph{reward}. A classic example of reinforcement learning is the grid world task, wherein the agent must navigate a maze: at each time-step, it perceives its current position, chooses to take a step in some direction, and, if this brings it to a goal that is located somewhere in the maze, receives a reward. This pattern of interactions fits in naturally with the structured learning environment outlined in section \ref{sec:learning_env}, with percepts specifying the setup and actions being the choice of a prediction. Only a small modification is required regarding rewards: if an agent is supposed to discover patterns and hidden variables by making predictions about the world, it should not rely on rewards provided by the environment, but instead be endowed with an internal mechanism by which it essentially rewards itself if the prediction was correct. 
(The idea of a learning process that does not primarily aim to achieve an 
externally supplied reward, but instead encourages a learner to explore its environment simply for the sake of obtaining more information (although that may turn out to be useful for reaching more external rewards in the future),
was incorporated from developmental psychology into reinforcement learning under the term \emph{intrinsically motivated learning} \citep{Oudeyer2007Intrinsic, Barto2013Intrinsically}. 
More specifically, intrisically motivated learning in RL often refers to mechanisms that guide the learner towards situations that maximise the gain of new information, which one might describe as curiosity. 
For the purpose of the present work, however, it is sufficient to consider an agent that simply rewards itself whenever it makes a correct prediction.)

\begin{figure}
\begin{centering}
\includegraphics[width=.45\textwidth]{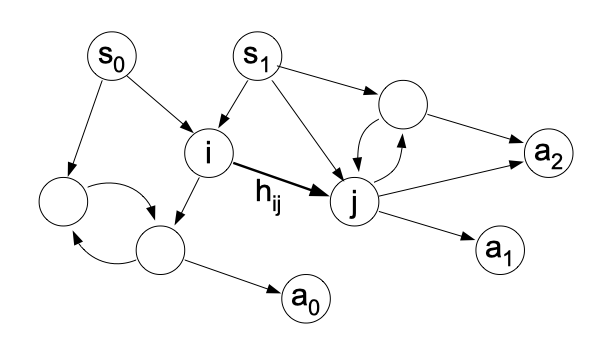}
\par\end{centering}
\caption{\label{fig:ECM_generic}
A PS agent's memory of its interaction with the environment is summarised in the \emph{episodic and compositional memory} (ECM): a network of \emph{clips} of previous experience, including in particular percepts (denoted $s_i$) and actions ($a_i$). Deliberation is realised as a random walk over clip space, starting at a percept and terminating at an action, with the probabilities of hopping from clip $i$ to clip $j$ governed by the weights $h_{ij}$ of the relevant edges. If a reward is received, the edges traversed to reach that decision are strengthened. 
}
\end{figure}

Let us now turn to the internal mechanism by which agents decide on an action given a percept, which is the defining feature for which projective simulation is named: PS agents simulate (or project) conceivable developments that, based on past experience, could arise from the present percept. Their simulation favours those sequences that have been rewarded in the past, so as to arrive at an action that is also likely to carry a reward. In order to ensure the autonomy and flexibility of the agent, the simulation is not based on some predefined representation of the environment, but instead on episodic `snippets' -- termed \emph{clips} -- from the agent's own experience, which could represent percepts, actions or combinations thereof. 
The deliberation process consists of a random walk over clip space, starting at the clip that represents the percept currently being presented and terminating when an action clip is reached and the corresponding action realised. A generic example of such a clip network is illustrated in Fig.~\ref{fig:ECM_generic}. 

In order to adapt its responses to an environment \textendash{} that is, to learn \textendash{} the agent must be able to modify how the random walk over clip space proceeds. To this end, each edge from clip $i$ to clip $j$ is given a (positive, real-valued) weight, termed the \emph{hopping value} or \emph{h-value} for short and denoted $h_{ij}$. These weights govern the probabilities with which the walk proceeds from clip $i$ to clip $j$:
\begin{equation}
P(j|i) = \frac{h_{ij}}{\sum_k h_{ik}},
\end{equation}
with all weights set initially to $h_{ij}=1$.
PS agents learn primarily by modifying the weights of edges: if a deliberation process going from percept $s$ to action $a$ leads to a reward $R$, then all the edges traversed as part of this deliberation are strengthened, i.e., their $h$-values are increased. In general, this is balanced by \emph{forgetting}, which decreases the weights of all edges by a factor $1-\gamma$, driving them back to their initial weight of $h_{ij}=1$, so as to gradually eliminate unused connections. Combining these two mechanisms, the update rule for $h$-values reads
\begin{equation}
h_{ij}^{\left(t+1\right)}-1=\left(1-\gamma\right)\left(h_{ij}^{\left(t\right)}-1\right)+\begin{cases}
R^{\left(t\right)} & \text{if used,}\\
0 & \text{if unused,}
\end{cases}\label{eq:PS update}
\end{equation}
where $R^{\left(t\right)}$ denotes the reward received at turn $t$.

The network of connected clips inside a PS agent is termed \emph{episodic and compositional memory} (ECM), based on two noteworthy properties: firstly, the sequence of clips that are excited during a random walk can be understood as a simulation of an ordered sequence of events, or \emph{episode}. Secondly, the set of clips over which the walk proceeds is not static, but can be augmented by creating new clips, either by \emph{composing} existing ones or by adding blank clips that can come to represent novel content. This second possibility, of additional clips that represent neither percepts nor actions, but some other, a priori undefined semantics, will enable our agents to form novel concepts. (While such clips can in principle be created dynamically, during the learning process, the present work focuses on how existing clips can acquire relevant semantics, leaving the exploration of clip creation to future work.)

\subsection{Enabling learning agents to handle more complex environments: connections to existing work}

\begin{figure}
\begin{centering}
\includegraphics[width=.85\textwidth]{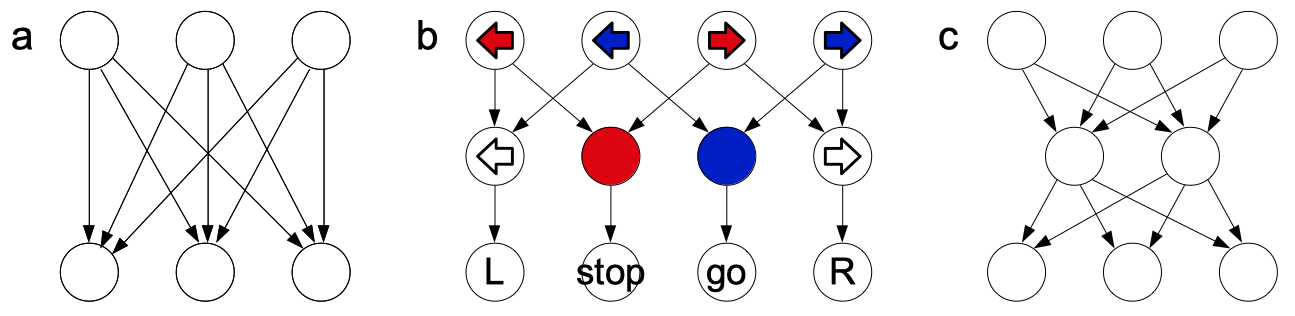}
\par\end{centering}
\caption{\label{fig:ECM_options}
Various ECM structures connecting percept clips (top layer) to action clips (bottom layer): (a) Two-layer is the simplest architecture. (b) Wildcard clips represent information about a subset of components of the percept. (c) 
Three-layer architecture proposed here, with a layer of intermediate clips that will come to represent values of hidden variables.}
\end{figure}

In the simplest agents, the ECM has just two layers, representing percepts and actions, with connections proceeding simply from percepts to actions, and their strengths encoding which is the preferred response to each input. Such a structure is shown in Fig.~\ref{fig:ECM_options}a. However, more complex tasks can generally be solved better with more sophisticated structures. By way of illustration, this section summarises a simple learning task that was previously posed to PS, and that resembles the abstraction task of the present work, before discussing previously proposed modifications that enable PS to handle this challenge. 

The task of interest is the infinite colour game, introduced by \cite{Melnikov2017Generalization}. In this environment, the agent is shown a two-component percept, featuring an arrow that points in a certain direction (left or right) and is painted in one of (countably) infinitely many colours. The agent then has the choice of moving left or right (ostensibly to defend one of two doors against an attacker) and is rewarded if it chose the correct action. The `hidden structure' in this environment is that the correct choice is telegraphed solely by the direction of the arrow, whereas the colour information is irrelevant to the task. The challenge for the agent is to learn to disregard colour, which would allow it to achieve perfect success in its responses even if it has never encountered a particular percept (that is, that combination of direction and colour) before. 

To solve this problem, \cite{Melnikov2017Generalization} introduced an architecture where the agent dynamically generates \emph{wildcard clips}: additional clips that are added to the ECM between the layers of percept and action clips, representing either only a direction without specifying a colour or only a colour without specifying a direction (or, most generally, neither a colour nor a direction, i.e., a completely uninformative clip). The structure is illustrated in Fig.~\ref{fig:ECM_options}b. Notably, the wildcard clips are connected to the two-component percept clips according to a fixed rule, namely connecting only to those percepts that contain the direction (resp. colour) in question. 
In order for such an agent to be successful in a given environment, the environment must have two key properties: the percept space must be formed by products of several components (or categories), which the agent must be able to perceive as independent pieces of information, and the reward rule must be such that disregarding a subset of these components is a useful strategy for determining the correct actions.
By capitalising on these properties, wildcard PS performed significantly better than chance on the infinite colour game. 
%
Here we want to abandon the assumption that the structure of the relevant variables is known a priori 
and aim instead to construct an agent that is able to \emph{infer} the structure of the variables from its interactions with the environment.  

Considering the above properties of the PS framework, we note that, while the architecture of PS agents supports reinforcement learning (RL), it differs from conventional, more widely used algorithms for RL or for machine learning in general in several important ways. Contrasting with RL algorithms such as SARSA or Q-Learning \citep{SuttonBarto1998}, which essentially tabulate the expected rewards for each percept-action pair, the additional internal structure of PS agents supports a more complex understanding of the environment. (In the case of environments that do not require such complexity, such as the classic benchmark tasks `grid world' and `mountain car', PS also achieves results that are at least comparable to conventional, tabular RL algorithms \citep{Melnikov2018Benchmarking}.) 

Another important and influential framework for machine learning employs artificial neural networks (ANNs). These networks also possess several layers or more convoluted structures and may, on first view, look quite similar to the ECM in Fig.~\ref{fig:ECM_options}c. However, the two structures function along very different lines. For one, training ANNs (by backpropagation) requires the learner to know what the outcome should have been (i.e., a supervised learning setting), whereas PS can learn by trial and error, i.e., being informed only whether it made the correct choice. With respect to the internal functioning, a single deliberative process in an ANN excites many neurons, often at the same time, with information being encoded in the \emph{pattern} of excitations, whereas individual neurons  typically carry no clear meaning. By contrast, in an ECM, exactly one clip is excited at a time, and any single clip can carry all the information involved in the deliberative process at that time (for example the entire percept or a complete specification of the action that the agent is deciding to take). This difference in functioning and semantics becomes especially relevant when one is concerned with interpretability: due to the delocalised way in which an ANN  represents information, it takes considerable effort to trace or explain how it reached the conclusion it did~\citep{Alvarez2018Interpretability, Sellam2019DeepBase, Biran2017Explanation, Molnar2019book, Samek2017Understanding}. PS, on the other hand, clearly reveals what path the deliberation took, passing through particular intermediate clips that can \textendash{} as we will see in the following \textendash{} be endowed with an objective interpretation in terms of hidden variables. 

\subsection{Specific architecture of our agents}
\label{sub:our_agent}

As we stressed in the previous section, the ambition of the present work is not simply that the agents learn to make correct predictions, but rather that they develop some internal representation of the hidden variables that underlie such predictions. At first blush, there is no obvious way to encode such representations in the simplest two-layer structure that is characteristic of the ECMs of standard simple PS agents. So, in order to support such representations, we propose an agent whose ECM consists of \emph{three} layers: an initial layer of percept clips (with one clip to represent each possible setup; in the default scenario, $27$), a final layer of action clips (representing the predictions for the various experiments, by default $18$), and, between them, a layer of \emph{intermediate clips} (denoted $\mc{I}$). 

We assume that these three layers are connected in a particular way, as illustrated in Fig.~\ref{fig:ECM_options}c.  (Note that the following specifies only which connections \emph{exist} in the ECM. The weights of the connections, on the other hand, which effectively guide the agent's choices and which will serve as a basis for identifying hidden variables, develop during the learning process.)
In our agents, each clip in one layer is connected to all the clips in the layer(s) immediately before and after, but not to any clips in the same layer or in more distant ones. Moreover, all connections are directed from percepts towards actions, so that the ECM is acyclic. Thus, every path from a percept clip to an action clip passes through exactly one intermediate clip on the way, and for every percept-action pair there is one path through each intermediate clip. 

Regarding the number of intermediate clips, we require only that it be no greater than the number of possible setups (percepts) the agent may encounter, and otherwise leave the number of clips in $\mc{I}$ unconstrained. This requirement is related to the natural interpretation of the intermediate clips. Intuitively, the idea is that each intermediate clip denotes a possible \emph{label} for a given situation (percept/setup). When the agent encounters a setup $s \in S$, they first have to choose a `label' for that experience. This is formalised as the random walk through the agent's ECM transitioning from $s$ to some $i \in \mc{I}$. Based on the label $i$, the agent then chooses an action \textendash{} formally, by transitioning from the intermediate clip $i$ to an action clip $a$.\footnote{For example, when looking at a traffic light an agent perceives a particular shade of green (determined by light and viewing conditions etc). They then disregard the particularities of that shade and simply label the experience as `green', and then go on to choose an action (e.g., driving) on the basis of that label.} 
Note that such labels may well be shared by various setups, but each meaningful label must be attached to at least one setup. For this reason, there is no point having more labels than there are setups to assign them to; hence the requirement that $|\mc{S}| \geq |\mc{I}|$. 
In the present work, we consider agents whose number of intermediate clips is equal to the number of hidden variables times the number of values that each variable can take. Preliminary tests suggest that having fewer intermediate clips than that is a significant obstacle to abstraction, whereas a larger number of clips leads to a slight reduction in learning efficiency, but does not pose any fundamental problems. We intend to explore this in more detail in future work. 


An additional feature of our proposed agents is a `boredom' mechanism, which addresses the following problem: once an agent has made the connection from a particular setup $s$ to one prediction $p$, which pertains to a particular experiment $e$, the most effective way for the agent to continue reaping rewards is to simply repeat prediction $p$ every time it encounters setup $s$. However, we want the agent to explore what would be the correct predictions for other experiments $e'$ as well. (The dilemma of balancing between these two goals is well-known in machine learning, where it is usually termed the `exploration vs exploitation' tradeoff.) To favour exploration, the agent is endowed with `boredom': if, for a give setup $s$ and a particular experiment $e$, the agent has come to favour one of the predictions that pertain to $e$ over the others with high probability, then experiment $e$ is deemed boring with this setup. Formally, for a given $s$, any prediction that pertains to an experiment that is deemed `boring' is rejected, with the deliberation process simply being reset until it produces a prediction about a non-boring experiment. We note that this rejection and resetting is an internal process applied by the agent itself. As far as the environment is concerned, the agent eventually produces a single prediction, which is guaranteed to pertain to an experiment that is not boring.

In order to highlight the capabilities that this architecture affords, we will compare the \emph{three-layer agents} described so far against simpler \emph{two-layer agents}, which lack an intermediate layer (see Fig.~\ref{fig:ECM_options}a). We will show that three-layer agents develop patterns of connection weights that can be interpreted as representing the environment's hidden variables and perform significantly better than chance on generalisation tests, whereas their two-layer counterparts are incapable of either of these feats.

\section{Variable Identification}
\label{sec:var_ident}

In the standard environment described in section \ref{sub:standard_scenario}, the agent is presented with an integer index specifying one of 27 possible setups, before subsequently choosing one of 18 available predictions (each of which pertains to one of 6 available experiments). The random walk leading to that decision consists of two steps: firstly, from the appropriate percept clip to one intermediate clip (which serves to `label' the given setup), and then onwards to an action clip representing a prediction. 
If the prediction is correct, then both of the connections traversed in the random walk will be strengthened in proportion to the agent's reward. Once this process has been iterated often enough, the agent should have learned both (i) to label each of the percepts $s$ with intermediate clips in $\mc{I}$, in the sense that the connections from $s$ to one or several particular $i$ are much stronger than to the others, and (ii) to choose correct outcome predictions for various experiments on the basis of those labels, in the sense that the connections from $i$ to some actions are much stronger than they are to others. 

Both of these sets of connections - from percepts to intermediate clips (representing assignments of labels to setups) and from intermediate clips to actions (encoding which labels are relevant to which experiments) - reflect patterns that the agent has learned in order to make sense of its environment. The present section provides a conceptual discussion of how certain properties and structures in the pattern of weights of these connections can be used to identify the abstract conceptual representations at play in the agent's deliberations. 

\subsection{Variable identification based on connections from percepts to intermediate clips} \label{sub:conceptual_percept_interm}

Before describing how we can identify the agent's abstract representations of the environment's hidden variables based on the weights of the connections in its ECM, it will be useful to specify more precisely what is meant by a `variable' in this context. Formally, a variable can be characterised as an abstract property such that every setup instantiates one and only one value 
%
%
of that property. This definition is trivially satisfied by the hidden variables `mass', `size' and `charge' in our running example. Importantly, this definition also implies that, for every variable, the set of values 
is jointly exhaustive and mutually exclusive with respect to setups, i.e., every setup maps to at least one value of a given variable, and no setup maps to more than one value of a given variable. 

\begin{figure}
\begin{centering}
\includegraphics[width=.8\textwidth]{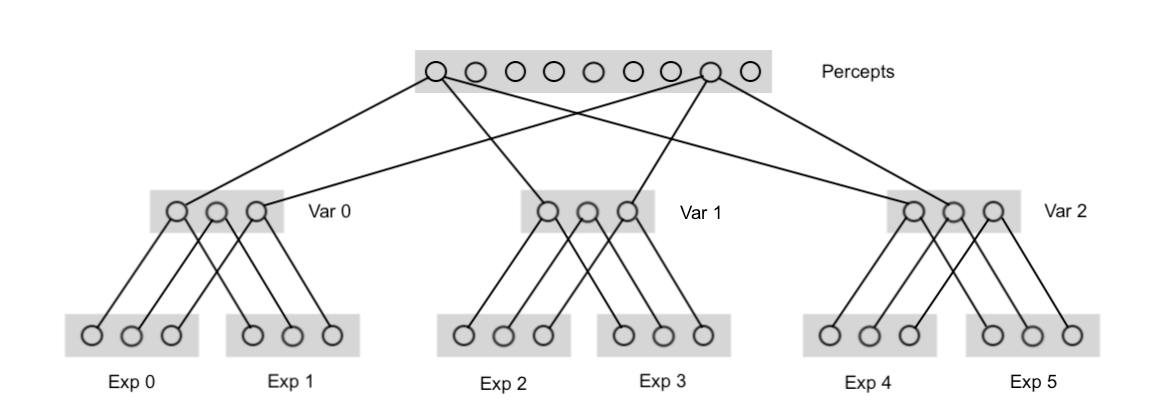}
\includegraphics[width=1.0\textwidth]{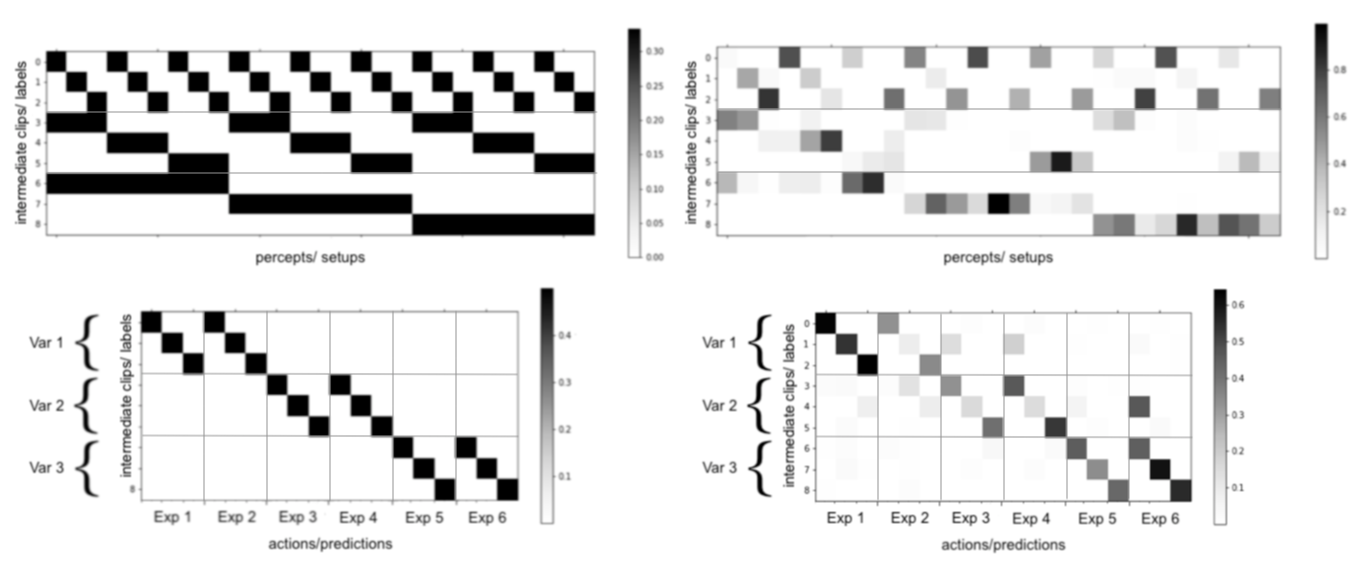}
\par\end{centering}
\caption{\label{fig:ECM_overview}
(top) Structure of connections in the ideal ECM. For clarity, only connections from two percepts are shown, and all weak connections are suppressed. (bottom) Matrix representations of the ideal connections (left) and, for comparison, the corresponding connections formed by a real agent (right), showing separately connections from percepts to intermediate clips (top row) and from intermediate clips to experiment-predictions (bottom row). In order to showcase the characteristic structure of those connections, the intermediate clips are ordered according to the values they represent, as $V_0=0$, $V_0=1$, $V_0=2$, $V_1=0$,  $V_1=1$, $V_1=2$, $V_2=0$, $V_2=1$, $V_2=2$. In the case of h-matrices learned by a real agent, the intermediate clips generally represent a random permutation of these values. However, they can be sorted in the same way by inferring the correct ordering based on an analysis of the connections, as detailed in section \ref{sec:results_main}.
}
\end{figure}

We can now detail what role such variables play in an agent's deliberation on a learning task. The most natural description is that the agent's deliberations about which prediction to make go by way of labelling the presented setup with a value of the relevant variable. To illustrate: the variable `size' plays a role in the agent's deliberations if and only if, in deciding which prediction to make for a given setup, they label the setup with a particular value for the `size' variable (e.g., `big', `small', `medium') and then choose the prediction on the basis of that label. This explication suggests that, when trying to identify the variables represented in the agent's deliberative structures, we should expect each value of a variable to be represented by a label for setups, i.e., by an intermediate clip in $\mc{I}$. Accordingly, a whole variable should be represented by a \emph{subset of intermediate clips}, denoted $\tilde{I}$, whose elements represent the various values of the given variable. 
Moreover, the sets representing different variables should be `mutually exclusive' and `jointly exhaustive' in the sense that for any setup $s$, the agent is disposed to label $s$ by exactly one of the labels in the set. Roughly, this means that each percept connects `strongly' to exactly one of the labels in the set, and `weakly' to all the other labels in the set. If two setups $s_{1}$ and $s_{2}$ both link strongly to different labels in the set representing a variable, that means that the setups have different values for that variable. If they link to the same label, they are perceived as sharing the same value for the variable. The top half of Fig.~\ref{fig:ECM_overview} illustrates the kind of pattern in the ECM that allows us to identify representations of variables via the semantics described above. 

In sum, then, the idea is this: in order to identify the abstract variables that are represented in the agent's deliberative structures, we should attempt to identify the subsets of intermediate clips in the agent's ECM that are mutually exclusive and jointly exhaustive with respect to setups. The functional role that these subsets play in the deliberations of a PS agent renders them susceptible to legitimate interpretation as internal representations of abstract variables. 

Having specified a criterion for identifying the variables at play in our agents' deliberations, we can now describe what it would look like for the agents to satisfy the central success criterion we put forward for the learning task described in section \ref{subsec:env_basic}: namely, that the agents form internal representations of the environment's hidden variables, i.e., `mass', `size' and `charge'. Following the procedure described above, we would identify these representations with subsets of labels (intermediate clips in $\mc{I}$) that are jointly exhaustive and mutually exclusive with respect to setups. Specifically, we would expect each of the three variables to be represented by a separate (pairwise disjoint\footnote{We expect these sets of intermediate clips to be pairwise disjoint because knowing, for a given setup, that a variable $V_0$ takes a value $v_0$ or that a variable $V_1$ takes a value $v_1$ are two independent pieces of information, or labels, that can be assigned to the setup.}) set of three intermediate clips (since each coarse-grained hidden variable has only three values) such that each of the 27 available setups is strongly connected to exactly one clip in each set. (One can verify that, for any clip in any set, there should therefore be 9 setups that map strongly to that clip, since for each value of charge/mass/size, there are 9 objects which have that value.) In the event that the agent forms a structure of this form in their ECM, we will be able to legitimately identify representations of the environment's three hidden variables in their internal deliberative structures. 

At this stage, it is worth pausing to reiterate a few important clarifications. Firstly, we stress that, in general (and in the specific example considered here), we assume that, for each hidden variable, there are several experiments whose outcomes are determined by the value that that variable takes for the given setup. If there were only one such experiment for a particular hidden variable, then the conceptual distinction between the hidden variable and the experiment that reveals it would be lost, and the intermediate clips would no longer represent abstractions, but would simply act as copies of the outcomes associated with the values of the hidden variables. The interpretation of intermediate clips as internal representations of the values of hidden variables is only principled and legitimate when the environment structure is rich enough to support abstraction, which in this case means that hidden variables are tested by multiple experiments. 

Secondly, we stress that, while the outcome of an experiment $M$ (`What happens when I hold the object next to a known magnet?') can be predicted by knowing the value of a hidden variable $\mu$ (`Is it magnetic?'), the two are conceptually very different objects. Crucially, the experiment $M$ is part of the agent's repertoire of actions, whereas $\mu$ is a hidden variable, i.e., a property of the environment that is in principle inaccessible to the agent, and whose existence and role the agent can only \emph{infer} from patterns in the way setups connect to (correct) predictions. 

\subsection{Variable identification based on connections from labels to experiment-predictions}

We turn now to presenting a second, alternative method of identifying the variables at play in the deliberations of our PS agents. We will see in section \ref{sec:results_main} that the two methods produce largely identical results.

If, as above, one wants to group the intermediate clips/labels into subsets such that each set represents the different values of a single variable, one could also simply pick one experiment and map backwards to the labels that predict its various outcomes. The resulting set of labels is then naturally interpreted as representing the variable tested by the given experiment. Ideally, there should be exactly one such label for each prediction, since we have assumed that each of the predictions associated with a given experiment correspond to values of the variable tested by that experiment. Moreover, if there exist experiments $e_1$, $e_2$ whose outcomes are predicted by the same variable, then one expects the sets of labels obtained in this manner to coincide. This allows one to verify that $e_1$ and $e_2$ are predicted by the same variable and, moreover, 
to identify which prediction of $e_1$ corresponds to the same value of the hidden variable as a particular prediction\footnote{In general, it may be that $e_1$ and $e_2$ reveal different coarse-grainings of a single variable, so that some pair of values of that variable lead to the same prediction in $e_1$ but different predictions in $e_2$, while another pair of values is distinguished only by $e_1$ but not by $e_2$. In this case, one cannot identify labels and predictions for both experiments one-to-one. However, and more importantly, it still holds that a single \emph{set} of labels connect strongly to all the predictions of $e_1$ and $e_2$, therefore still supporting the inference that these labels collectively represent a single variable that predicts the outcomes of both experiments. We have deliberately excluded such differently coarse-grained experiments in the scenario considered here in order to focus on more essential questions.} for $e_2$. On the other hand, if two experiments $e_1$ and $e_3$ are predicted by different variables, then one should expect that any label that is strongly connected to a prediction of $e_1$ is not strongly connected to any prediction pertaining to $e_3$. The expected pattern of connections is illustrated in the bottom half of Fig.~\ref{fig:ECM_overview}.

In sum, the idea is that one can identify the different values of a single variable by identifying those labels that lead to all the different predictions of a single experiment. If there are two experiments whose various predictions are reached from the same set of labels, then these should be interpreted as being predicted by the same variable, whereas disjoint sets of labels herald experiments that reveal different variables. 

Again, we can illustrate this second prospective semantics for identifying representations of hidden variables in the agent's ECM by considering the example presented in section \ref{sub:standard_scenario}. As before, the aim is that the agent form internal representations of the coarse-grained variables we interpret as `mass', `size' and `charge'. The new procedure for identifying these representations works as follows. For each experiment, we check whether there is a set of intermediate clips such that every clip in the set connects strongly to a different possible prediction for that experiment. For example, in the experiment in which the given object is placed on a scale, we check whether there exists a set $\tilde{I} \subseteq \mc{I}$ such that each $i \in \tilde{I}$ connects strongly to one of the three possible predictions for the experiment (`high reading', `low reading', `medium reading'). If such a $\tilde{I}$ exists, then we can interpret $\tilde{I}$ as the agent's internal representation of a variable that predicts the outcome of that experiment. Moreover, we expect it to be that (i) each of the clips in $\tilde{I}$ also connect strongly to exactly one prediction of the other mass experiment, and (ii) none of the clips in $\tilde{I}$ connect strongly to any of the predictions associated with any of the size or charge experiments. This fact allows the agent to deduce that there exists a single variable that predicts the outcomes of both the `scale' and the `momentum' experiment, but not the others. We, human scientists, might subsequently identify this variable as `mass', but the essential inference that there exists such a variable can be made by the agent itself. 

Finally, let us preempt a potential criticism that one might raise against this second procedure for identifying representations of variables in the agents' ECM. Specifically, one might argue that by assuming that the agent knows that the number of values of each variable should correspond to the number of outcomes of some available experiment, we are essentially giving them a-priori knowledge about the hidden structure of their environment, and thereby trivialising the discovery task. However, we hold that, firstly, the agent can make the non-trivial inference that there exists an unobserved variable whose value predicts the outcomes of one or more experiments. Moreover, the agent learns to distinguish between several coarse-grained \emph{intervals} of values that this variable can take that map to different predictions in the experiments. The semantics we are proposing makes no ontological claims about the values that the unobserved variable itself takes, but simply points out that there exist patterns in the environment that can be explained in terms of hidden variables. This is the essential insight that the agent distills, and it does not depend on any a-priori assumptions about the number of values this variable might take. 



\section{Results}
\label{sec:results_main}

\begin{figure}
\begin{centering}
\includegraphics[width=0.4\textwidth]{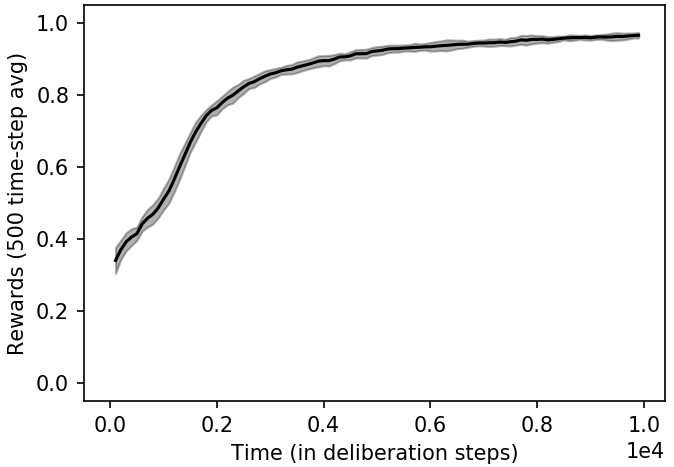}
\includegraphics[width=0.4\textwidth]{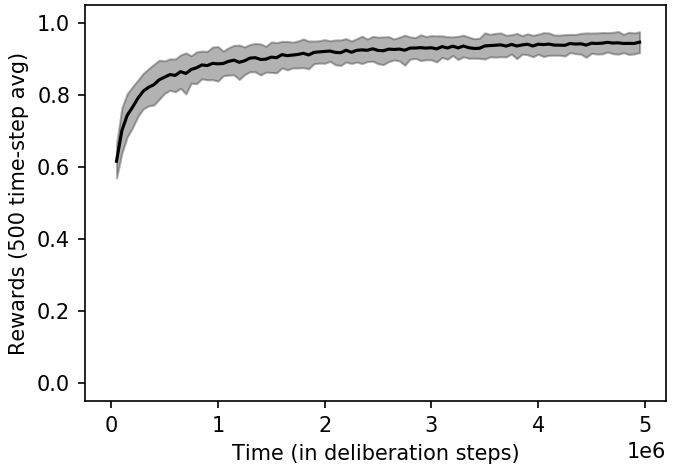}
\par\end{centering}
\caption{\label{fig:learning_curve}
Comparison of the reward rate as a function of time achieved by simple two-layer agents, which are incapable of abstraction, and three-layer agents, which are capable of abstraction. Note the different time-scales required to achieve rewards around 90\%: two-layer agents were trained for only $10^4$ rounds of interaction, while three-layer agents were given $T=5*10^6$.}
\end{figure}

The first result of our simulations is that our three-layer agents learn to successfully predict the outcomes of setup-experiment pairs with success probabilities of at least $90\%$. One can compare how quickly the three-layer agent proposed here learns compared to a basic two-layer agent that simply tabulates the correct prediction for each percept-experiment pair. As shown in Fig.~\ref{fig:learning_curve}, two-layer agents learn much more quickly.

It should not come as a surprise that, in standard reinforcement learning tasks, three-layer agents are much slower when it comes to learning how to maximise rewards, since the extra clip-layer significantly complicates their deliberations and interferes with their memorisation of previous rewards. However, given that we are interested in engineering agents which are capable not just of maximising rewards, but also of forming interpretable conceptual representations of their environment, we do not take this to constitute a major problem. It is utopian to expect that one could satisfy the second of these success criteria without sacrificing something in the way of learning speed. But while they incur operational disadvantages regarding learning speed, we will see that three-layer agents accrue some major operational advantages pertaining to their ability to solve generalisation problems (see section \ref{sub:results_gen}). 

While establishing that agents learn how to make correct predictions and maximise rewards is important, the main point we want to make in this section is that agents with the three-layer ECM structure outlined above do indeed develop identifiable abstract representations of the environment's hidden variables, i.e., they satisfy the central success criterion of the present work. To justify this claim, the following subsections detail firstly an analysis of the connections that the agent establishes between percepts and intermediate clips and how those represent abstractions and allow us to identify which subsets of intermediate clips represent variables, followed by an analogous analysis based on the connections from intermediate clips to actions (predictions). Finally, section \ref{sub:results_gen} turns to the problem of generalisation and demonstrates that, while two-layer agents are constitutionally incapable of solving the task (or of forming meaningful abstractions), our three-layer agents achieve a significantly better performance.

\subsection{Verifying abstraction and identifying variables based on connections between percepts and intermediate clips}

One way of analysing what the agent has learned is based on the conceptual considerations laid out in section \ref{sub:conceptual_percept_interm}. We formalise the requirements of exhaustivity and exclusivity as follows: given a subset of intermediate clips $\tilde{I} \subseteq I$ that might represent (the set of values of) a hidden variable, we define functions $exh(\tilde{I})$ and $excl(\tilde{I})$ that assign to $\tilde{I}$ one real-valued indicator each, quantifying how well it satisfies exhaustivity and exclusivity, respectively. Intuitively, high values of $exh(\tilde{I})$ and $excl(\tilde{I})$ indicate that the elements of $\tilde{I}$ plausibly represent the values of a single variable identified by the agent. 

Exhaustivity demands that each percept $s$ be strongly connected to (at least) one clip in $\tilde{I}$. The condition is therefore violated, for a given $s$, if the probability of reaching any clip in $\tilde{I}$ -- technically, we take $\max_{i\in \tilde{I}} P(i|s)$ -- is much smaller than the probability of going to a clip outside the subset, which we quantify\footnote{In the case of the full set, $\tilde{I}=I$, this probability is zero.} by $\max_{i\in I\backslash \tilde{I}} P(i|s)$. As a measure of exhaustivity, we take a (weighted, logarithmic) average of the ratio of these probabilities over all percepts, 
\begin{equation}
exh(\tilde{I}) := \sum_{s} w_s \log\left(
\frac{\max\limits_{i\in \tilde{I}} P(i|s) }
{\max\limits_{i\in I\backslash \tilde{I}} P(i|s)} \right),
\end{equation} 
where $w_s$ is a vector of weights\footnote{For the data presented here, the weights $w_s$ for a given intermediate clip $i$ are determined as follows: the percepts $s$ are sorted according to the values $P(i|s)$ and indexed with integers $ind_s=0,1,2,...S-1$ such that smaller $P(i|s)$ carry larger indices. Then $w_s=(ind_s/(S-1))^3/N$, normalised to unit sum by choosing $N=\sum_{s}(ind_s/(S-1))^3$. This assigns higher weights (up to $1/N$) to those percepts that have small transition probabilities, effectively highlighting possible violations of the exhaustivity condition.}.
In an ideal agent and for subsets $\tilde{I}$ that actually represent a hidden variable, this measure is zero. Larger values can occur if the agent is more likely to go to clips inside $\tilde{I}$ than to any clips outside it, but, more importantly, values $<0$ herald a violation of exhaustivity.

Exclusivity demands that each percept $s$ map strongly to no more than one intermediate clip in $\tilde{I}$. The condition is therefore violated, for a given $s$, if the second-largest probability of reaching a clip in $\tilde{I}$ is comparable to the largest one. As a measure of exclusivity, we take the (weighted, logarithmic) average of the ratio of these probabilities over all percepts,
\begin{equation}
excl(\tilde{I}) := \sum_{s} w_s \log\left( \frac{\max\limits_{i\in \tilde{I}} P(i|s) }
{\text{sec}\max\limits_{i\in \tilde{I}} P(i|s)} \right),
\end{equation}
with the same weights $w_s$ as above. In an ideal agent and for subsets $\tilde{I}$ that actually represent a hidden variable, this measure tends to plus infinity, whereas values close to $1$ herald a violation of exclusivity. (The measure is non-negative by design.)

\begin{figure}
\begin{centering}
\includegraphics[width=0.7\textwidth]{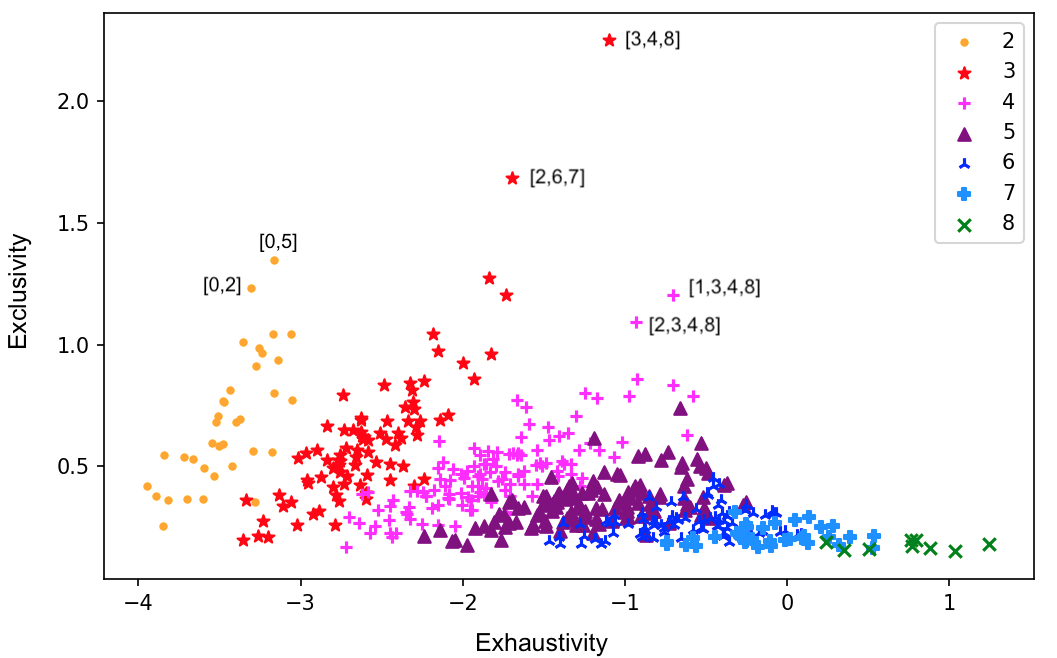}
\par\end{centering}
\caption{\label{fig:scatterplot}
Measures of exclusivity and exhaustivity for all subsets of intermediate clips, for the agent specified in section \ref{sub:our_agent} , with different symbols indicating the cardinality of the subset. Note how large subsets achieve comparatively high exhaustivity, but at the cost of violating exclusivity, whereas subsets of small cardinality have low exhaustivity but high exclusivity. Some of the subsets that achieve the highest values for both measures simultaneously are specified explicitly.}
\end{figure}

Any subset $\tilde{I}$ that is close to representing (the values of) a hidden variable must have large values of both exhaustivity and exclusivity. To check which $\tilde{I}$ satisfy this condition, we plot the two measures for all subsets of the set of intermediate clips in Fig.~\ref{fig:scatterplot}. 
Based on this analysis, one can identify a few `good' subsets; for example, in the particular agent analysed here, intermediate clips [3,4,8] are likely to represent one variable, while [2,6,7] are likely to represent another variable. 

In addition to identifying particular subsets of intermediate clips, this analysis also reveals, for example, how many hidden intermediate clips are necessary to represent a hidden variable exhaustively (given by the cardinality of the `good' subsets).
For an ensemble of 20 agents in the standard setting described in section \ref{sub:standard_scenario}, we obtain a value of $3.03\pm0.19$, clearly revealing that the environment, in fact, contains hidden variables that take 3 distinct values each.
A similar analysis can be performed based on the second layer of connections, as will be discussed in the following section. Fig.~\ref{fig:h2analys_compare} summarizes the results of this analysis and demonstrates how they allow one to read off essential parameters of the environment (in particular the number of values that the hidden variables can take) by tracking how the results change across different environments.

\begin{figure}
\begin{centering}
\includegraphics[width=1.\textwidth]{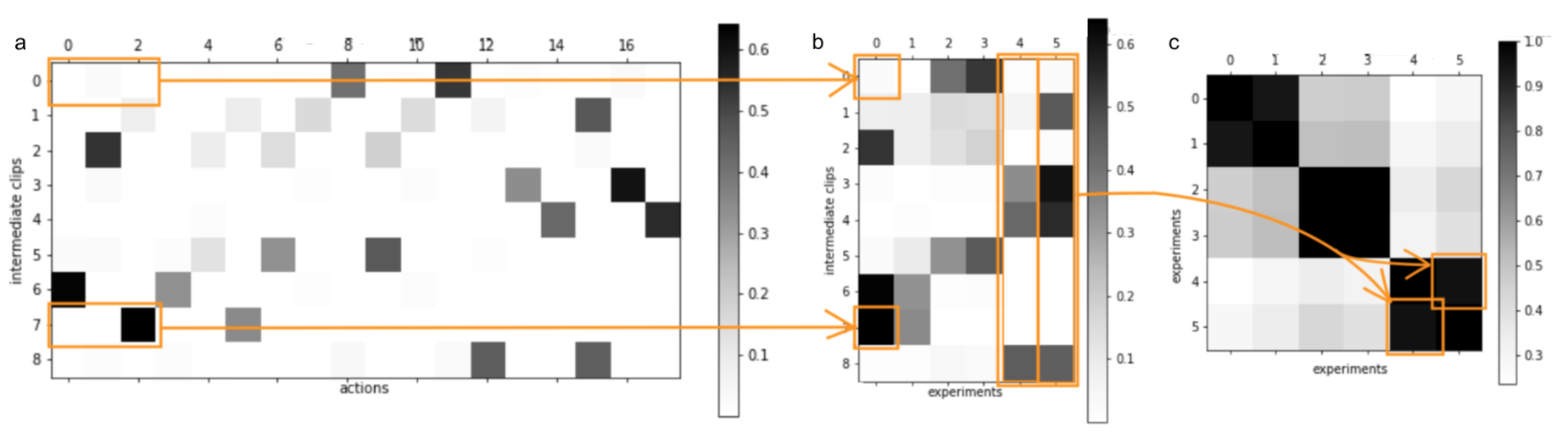}
\par\end{centering}
\caption{\label{fig:h2analys}
Identifying variables based on the connections from intermediate to action clips. (a) Transition probabilities from intermediate clips to predictions for each experiment $e$. These allow one to compute with how much certainty (quantified by the neg-entropy) each intermediate clip predicts the outcome of each experiment. (b) In a table of how well each intermediate clip predicts the outcomes of each experiment, comparing two columns (the  `predictability profiles') of two experiments allows one to judge how likely they are to involve the same variable. (c) The table of `predictability correlations' between experiments has a striking block-diagonal structure, clearly showing that experiments 0 and 1 are predicted by one variable, 2 and 3 by another and 4 and 5 by a third. 
(Note that the suggestive ordering of pairs of correlated experiments in panel c is due to the way the environment was coded in our simulations. However, the high contrast of the correlation matrix allows one to identify related experiments in generic environments that do not have this ordering just as well. Note also that the correlation matrix is symmetric under transposition by construction, since the measure of correlation is independent of the order of the experiments being compared.)
Working backwards, one can identify in panel (b) that, for example, experiments 4 and 5 are predicted most prominently by intermediate clips [3,4,8], and one can further verify in panel (a) that those intermediate clips represent different values of the underlying variable, since they lead to mutually exclusive and jointly exhaustive predictions for the experiments in question.}
\end{figure}

\subsection{Verifying abstraction and identifying variables based on connections between intermediate and action clips}

The procedure for analysing the connections from intermediate to action clips is summarised in in Fig.~\ref{fig:h2analys}. One begins by quantifying how strongly each intermediate clip predicts the outcomes of each experiment, which allows one to group experiments whose outcomes are predicted by the same subsets of intermediate clips together. Each such group is considered to stem from one hidden variable. Working backwards, one can then identify which intermediate clips represent values of each variable.
This analysis reveals how many hidden variables are necessary to predict the outcomes of all experiments in question, and moreover how many - and, in fact, which - experiments are predicted by each of those variables. As for the intermediate clips, one can identify which intermediate clips represent the various values of each of those variables. 

For example, for the individual agent analysed here, the analysis identifies experiments 0 and 1 as being predicted by one variable, whose values are best represented by intermediate clips [2,6,7]; similarly 2 and 3 are predicted by a hidden variable whose values are represented by intermediate clips [0,1,5], and experiments 4 and 5 are predicted by intermediate clips [3,4,8].
Let us compare this conclusion with the analysis based on the connections from percepts to intermediate clips, shown in Fig.~\ref{fig:scatterplot}: notably, while set set [0,1,5] was not highlighted in Fig.~\ref{fig:scatterplot}, the two subsets that are identified most clearly in Fig.~\ref{fig:scatterplot}, [2,6,7] and [3,4,8], are the same ones found in the present analysis. 

\begin{figure}
\begin{centering}
\includegraphics[width=0.3\textwidth]{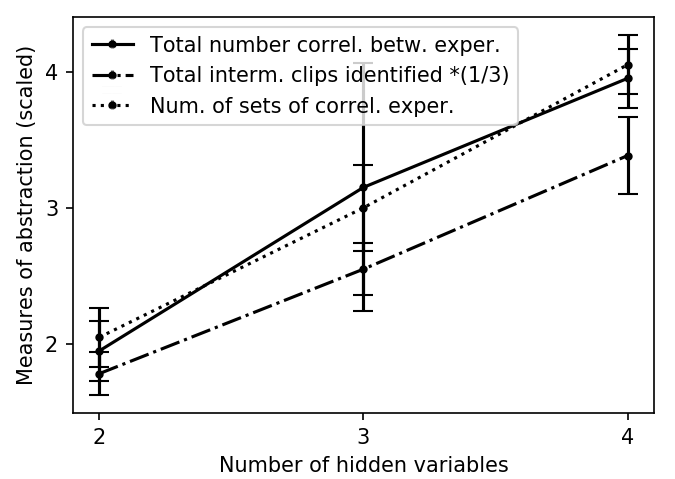}
\includegraphics[width=0.3\textwidth]{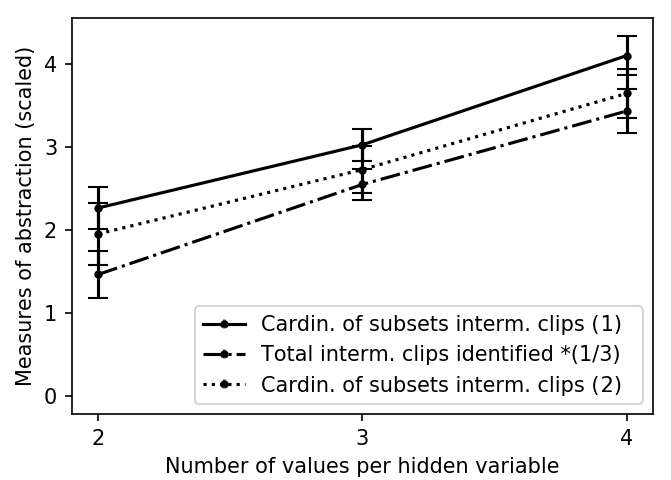}
\includegraphics[width=0.3\textwidth]{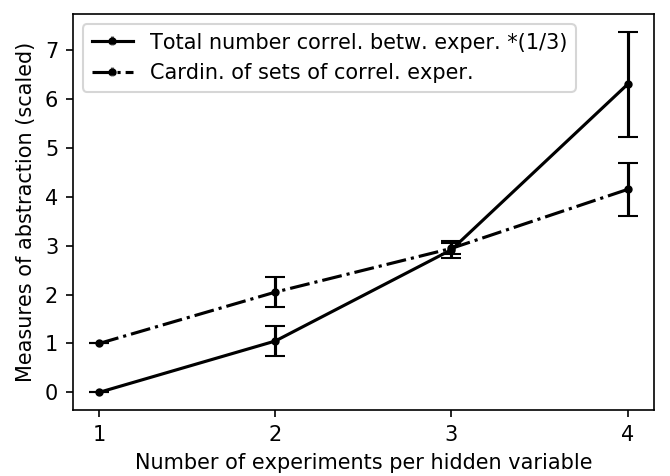}
\par\end{centering}
\caption{\label{fig:h2analys_compare}
Identifying properties of the environment based on an analysis of the connections between percepts, intermediate clips and actions established by the agent (h-matrix): 
(a) One can read off the number of hidden variables $|\mc{V}|$ by determining the number of disjoint blocks in the correlation matrix depicted in Fig.~\ref{fig:h2analys}c, or, indirectly, from the number of intermediate clips identified as representing values of variables (which should be $|\mc{V}|*|\mc{O}|$) or from the number of non-trivial correlations between experiments (which should be $|\mc{V}|*|\mc{E}|(|\mc{E}|-1)/2$).
(b) One can read off the number of values each hidden variable can take from the cardinality of the sets of intermediate clips identified as representing single variables (based on either the connections to percepts (1) or the connections to actions (2)), or, indirectly, from the number of intermediate clips identified as representing values of variables (which should be $|\mc{V}|*|\mc{O}|$).
(c) One can read off the number of experiments whose outcomes are predicted by each hidden variable by determining directly how many experiments are strongly correlated in the matrix depicted in Fig.~\ref{fig:h2analys}c, or, indirectly, from the number of non-trivial correlations between experiments (which should be $|\mc{V}|*|\mc{E}|(|\mc{E}|-1)/2$).
Note that agents in different environments were trained for different durations $T$ (measured in interaction rounds), with the training times for each environment chosen such that the agents' h-matrices settled into a clear pattern, as shown by the fact that the various measures used for analysing the abstractions formed by the agent no longer changed noticeably.
Specifically, agents in the default scenario ($(|\mc{V}|,|\tilde{\mc{E}}|,|\mc{O}|)=(3,2,3)$) were trained for 
$T=5*10^6$ time-steps, whereas environments with different values used (a) ($|\mc{V}|=2$, $T=5*10^5$), ($|\mc{V}|=4$, $T=5*10^7$), ($|\mc{V}|=5$, $T=10^8$), (b) ($|\tilde{\mc{E}}|=1$, $T=5*10^5$), ($|\tilde{\mc{E}}|=3$, $T=5*10^6$), ($|\tilde{\mc{E}}|=4$, $T=5*10^6$), and (c) ($|\mc{O}|=2$, $T=5*10^5$), ($|\mc{O}|=4$, $T=10^7$).
}
\end{figure}

Regarding parameters of the environment, for an ensemble of 20 agents in the standard setting described in section \ref{sub:standard_scenario}, we obtain the following measures:
\begin{itemize}
\item number of hidden variables (distinct classes of correlated experiments): $3.00\pm0.32$
\item number of experiments predicted by each variable (number of experiments with which each experiment correlates strongly): $2.05\pm0.30$
\item number of values each variable can take (cardinality of the sets of intermediate variables identified): $2.73\pm0.29$
\item number of distinct intermediate clips identified as best representatives of values of hidden variables: $7.65\pm0.57$ (this should be num\_features*num\_values) 
\end{itemize}
Fig.~\ref{fig:h2analys_compare} shows how these measures change across environments with different numbers of hidden variables, of values and of experiments per variable.

\subsection{Generalisation}
\label{sub:results_gen}

\begin{figure}
\begin{centering}
\includegraphics[width=0.4\textwidth]{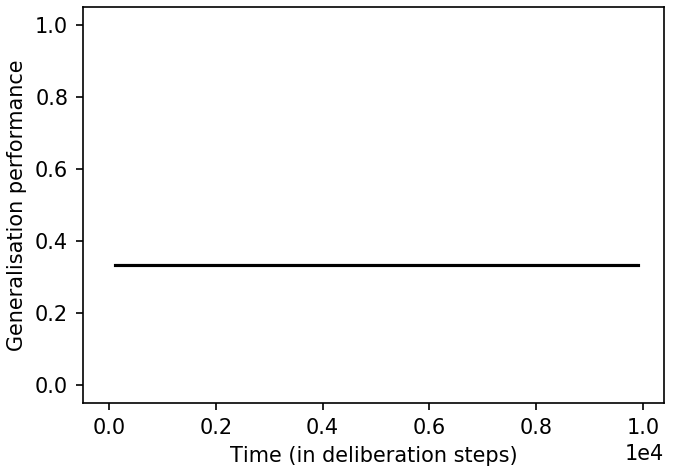}
\includegraphics[width=0.4\textwidth]{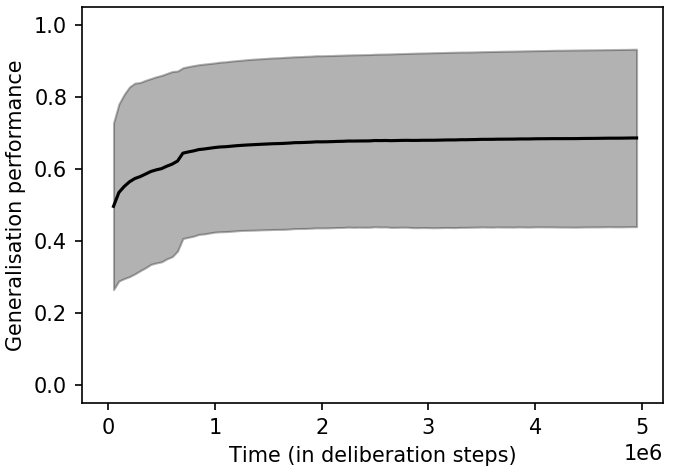}
\par\end{centering}
\caption{\label{fig:generalisation}
Performance on generalisation task over the course of training: (a) two-layer agents never go beyond chance level (1/number of possible outcomes, i.e., $1/3$), whereas (b) three-layer agents achieve significantly higher success probability; in fact on par with their success rate at the percept-experiment pairs for which they did receive feedback.}
\end{figure}

To illustrate that the agent can, in fact, reap operational benefits from this construction, consider a problem where an agent only trains with a subset of objects and experiments, leaving out one object-experiment pair, but is then tested on the pair that it has never encountered. (In order to allow repeated testing at different stages of the learning process, these agents never receive feedback on the `test' task.) Fig.~\ref{fig:generalisation} illustrates how
two-layer agents can only guess at random in that case,
whereas an ensemble of our three-layer agents achieve significantly higher reward rates (on the validation test), of $(69\pm25)\%$.
 This provides a concrete empirical vindication of the conjecture that the cognitive faculties of abstraction and generalisation are intimately related. 

\section{Discussion}
\label{sec:discussion}

Before concluding, we pause to discuss and reiterate some of the major implications of the preceding results and analysis. 

\subsection{On the Legitimacy of the Prospective `Concepts'}

In sections \ref{sec:var_ident} and \ref{sec:results_main}, we presented an operational semantics for identifying the conceptual representations at play in the deliberations of PS agents, before empirically illustrating how these representations are formed and identified in concrete learning tasks. At this stage, one might be inclined to dispute the extent to which the identified representations are truly indicative of underlying conceptual thought. While we acknowledge that the defining characteristics of true conceptual thought are a topic of substantial ongoing philosophical debate and greatly exceed the ambit of the present work, it is worth highlighting that the semantics for concept identification presented here chimes well with mainstream philosophical views about the nature of conceptual thought. 

Specifically, there is an influential view according to which the defining hallmark of conceptual thought is \emph{compositionality}. On this view, an agent can only be said to truly possess a concept if they are able to use it in a completely general way that is not restricted to any particular cognitive context. For example, an agent could not truly be said to posses the concept `red' if the only thought in which they are able to use the concept is `big red car'. In order to qualify as truly possessing the concept, the agent should be able to separate it from that one thought and use it to construct new thoughts like `small red van' or `sparkly red necklace'.  This idea was formally codified in Evans' (1982) so called `generality constraint', which posits that in order to qualify as engaging in genuine conceptual thought, an agent must be capable of entertaining all syntactically permissible combinations of the concepts they purportedly posses. So if the agent purportedly possesses the concepts `red', `blue', `heavy', `van', `house' and `guitar', they should be able to form the thoughts `red van', `red house', `red guitar', `blue van', `blue house', `blue guitar', `heavy van', `heavy house' and `heavy guitar' (but not, e.g., `red blue' or `house guitar', which are not syntactically permissible). Variations of this generality constraint have been used to evaluate the prospective possession of genuine concepts by various animals. For example, \cite{Carruthers2009Invertebrate} argues that the Australian digger wasp, which uses its body to measure the length of the sand towers it constructs, does not employ the concept of `length' in a sufficiently compositional manner to warrant ascribing possession of the concept to it. 

At this stage, it is pertinent to ask whether our ascription of concepts representing hidden variables to PS agents is consistent with Evans' generality constraint or the requirement of compositionality more generally. Happily, it seems that the answer to this question is at least partially positive. To see this, recall that on our semantics, one of the central criteria for a set of labels to represent a variable is that they be jointly exhaustive with respect to setups. This means that for every object that the agent is capable of representing, they are strongly disposed to label that object with one of the values of the relevant variable. This in turn implies that each value of the identified variable will connect strongly to multiple objects. If we identify concepts with values of variables,\footnote{So that the concept `red' corresponds to one possible value of the colour variable, for example.} then this means that the concepts formed by the agent are always applicable to multiple objects and therefore exhibit at least a moderate degree of compositionality. Although this is not enough to guarantee full accordance with Evans' generality constraint, which demands complete and unrestricted compositionality, it is similar to the kind of partial compositionality that authors like Carruthers (2009) argue is sufficient for genuine conceptual thought. 

\subsection{From Abstraction to Generalisation}

The majority of our discussion so far has focused on \emph{abstraction}, i.e., the capacity to form abstract conceptual representations of salient features of one's environment. A closely related (but importantly distinct) phenomenon is \emph{generalisation}, meaning the capacity to utilise the knowledge that one has acquired through previous experiences to deal efficiently with \emph{new} experiences that differ from everything  one has previously encountered. It is natural to conjecture that an agent's ability to solve generalisation problems is closely related to their ability to solve abstraction problems. One of the major payoffs of the present analysis is that we are able to provide a concrete empirical vindication of this conjecture, by showing that agents that form identifiable variable representations (in the sense described above) are better able to solve generalisation tasks than agents that lack the cognitive capacity to form these representations (see section \ref{sub:results_gen}). 

We also noted that this power comes at a cost, with simple two-layer agents reaching high rates of correct predictions much more quickly than the more sophisticated three-layer agents (see Fig.~\ref{fig:learning_curve}). In an environment where memorisation of percept-action pairs is a viable strategy, it may therefore be most efficient to employ a two-layer agent, which does not waste time looking for hidden variables. However, as one proceeds to larger, more complex environments, where the agent will more frequently encounter percepts it has not seen before, the ability to generalise (in particular by forming abstractions) becomes increasingly advantageous. 

The task of facilitating meaningful generalisation in PS agents has been the focus of previous work, most notably by \cite{Melnikov2017Generalization}. The generalisation capabilities exhibited by the agents considered in section \ref{sub:results_gen} go significantly beyond anything in this existing literature. Most importantly, our agents are able to successfully abstract and generalise in a way that does not rely on equipping them with a-priori knowledge regarding the structure of the environment. In contrast, the generalisation mechanisms described by \citeauthor{Melnikov2017Generalization} rely on learning rules which implicitly encode a priori knowledge regarding the way in which the percept space can be coded by values of the environment's hidden variables. In our framework, the agent discovers this hidden variable structure for themselves, and the very act of doing so facilitates their ability to generalise. No extra learning rule is required. 

It is also instructive here to consider the relationship between abstraction and generalisation in the context of neural network architectures, which are of course remarkably successful in a wide array of practical generalisation problems that involve generalising the patterns encountered in the training set to deal with novel data in the test set. Typically, the networks are able to achieve this generalisation capability without developing any easily identifiable representations of abstract concepts. 
This suggests that, while abstraction can be a helpful basis for generalisation, as we have argued in the present work, it is not a necessary pre-requisite. On the other hand, the ability of artificial neural networks to generalise from their training examples to test instances has recently been cast into doubt, with the appearance of striking results of adversarial approaches: notably, \cite{Moosavi2015DeepFool} proposed an algorithm that systematically fools deep neural networks into misclassifying images by manipulating just a few pixels. Such results cast serious doubt on the reliability of broad and deep neural networks and highlight the importance of transparency in building more robust ML solutions. 

In this context, it is worth discussing the work of \cite{Iten2018Discovering}, who trained neural networks in such a way that they managed to `extract simple physical concepts from experimental data'. A fundamental component of this process is an autoencoder, which is trained to compress input data (such as time-series data from a damped pendulum) through a bottleneck of just a few so-called latent neurons before attempting to reconstruct the original input. By testing how well 
the network can later reconstruct the given data or make predictions based on it,
given a varying number of neurons in this bottleneck, one can infer how many real parameters are needed to specify a particular instance from among the family of inputs on which the autoencoder was trained. For example, if one is drawing from a family of time-series data for damped pendula, the parameters that specify one instance are the frequency, damping parameter, initial phase and amplitude. The values that these parameters take can be recovered from the excitations of the latent neurons in the autoencoder. 

An obvious difference between \cite{Iten2018Discovering} and the present work is the implementation that supports the learning process (artificial neural networks in one case, projective simulation in the other). However, a more interesting point for the present discussion are the conceptual differences regarding \emph{what} is learned in each case, rather than how.
One fundamental difference is that we consider agents that \emph{explore} their environment by interacting with it and, accordingly, adopt the paradigm of reinforcement learning. By contrast, continuing with the example of damped oscillators, \citeauthor{Iten2018Discovering} consider an algorithm that is fed pre-recorded data - one might imagine being given a notebook with observations made in a laboratory, but no opportunity to go to the lab and experiment oneself. While learning from pre-recorded data is a powerful paradigm that has achieved great success for certain classes of problems, it requires the implicit assumption that there was already some entity that gathered the data, and, more fundamentally, that \emph{identified relevant variables} whose values should be recorded for subsequent analysis. 

It is this pre-requisite for data-based learning that our agents address: they start from a setting where it is not known how a stream of complex sensory input should be decomposed into independent, meaningful variables. 
This problem is not as far-fetched as one might think: in the early development of various theories, for example quantum mechanics and electromagnetism, it was a point of considerable debate which variables or concepts might be useful in talking about the subject, and progress was only made by experimentation - that is, by interacting with the systems under study. In our formal framework, this absence of pre-existing variables is reflected in the fact that we consider percepts as being labelled by unique, atomic indices rather than vectors consisting of well-defined components. Our agents take the first basic step of classifying these percepts by ascribing to them operationally meaningful labels, which, crucially, have a particular structure, with groups of labels forming a mutually exclusive and jointly exhaustive classification of percepts. We argue that this property of a set of labels is the defining feature that allows one to interpret them as representing values of some unobserved variable. In this sense, our agents can discover the existence of hidden environmental variables.

The question of how one might infer the values of such variables from the available perceptual data is a second, distinct step in learning about the environment. Our agents, facing an environment that is less challenging in this regard, can essentially memorise the value of each variable for each percept. \citeauthor{Iten2018Discovering} offer a more sophisticated approach to this part of the problem, implicitly modelling the relation between the new-found variables by learning to compress families of curves relating their values. However, we note that such compression can only be successful if one ensures that all curves are drawn from the same family (for example, recording the position over time for damped oscillators). In order to ensure that each data-set instantiates the relation between the same pair (or set) of variables and that other relevant circumstances are kept constant throughout, one must once again first identify the relevant variables for the system under study. 


It is the ability to perform this first, more fundamental step, of autonomously discovering that unstructured, atomic percepts admit a decomposition into meaningful variables, that is missing in the aforementioned examples using neural networks. It seems plausible that this ability might support a more robust performance, in particular with regards to generalisation.


\subsection{Transparency, Explanation and Abstraction}

As well as allowing PS agents to accrue significant new operational capacities in generalisation tasks, the ability to form abstract conceptual representations also promises a number of other advantages. One of the most salient advantages relates to the problem of rendering the deliberations and decisions of PS agents fully communicable, explicable and transparent, a problem that becomes urgent whenever artificial intelligence is put to practical use in human society. 

To see this, note first that the present work takes the first steps towards constructing an explicit symbolic interface through which PS agents can naturally articulate and communicate explanations of their reasoning processes and decisions. For example, once the semantics has been employed to identify the variables corresponding to `mass', `size' and `charge' in the agent's deliberations, it would be straightforward to implement an automatic explanation generator that provided explicit linguistic explanations of all the agent's actions, e.g., `I predicted that the scale reading would be high because object 2 is heavy'. Although the exact definition of agent transparency is still a matter of significant controversy in the current literature (see, e.g., \cite{Chen2014Situation, Lyons2014Transparency}), it seems clear that the ability to automatically construct explicit explanations of an agent's actions and deliberations constitutes a major step towards `transparency' on all plausible interpretations of the term. For example, \citeauthor{Chen2014Situation} define agent transparency as `the quality of an interface (e.g., visual, linguistic) pertaining to its abilities to afford...comprehension about an intelligent agent's intent, performance, future plans, and reasoning process'. It is obvious that the present work makes significant strides towards bringing PS agents in line with this criterion. 




\section{Future Work and Conclusion}

Finally, we conclude by highlighting promising avenues to be explored in future work. 

The first avenue relates to one of the most distinctive and crucial cognitive capacities of human reasoners, namely the ability to identify and exploit correlations between variables in their environment. Here we have addressed one fundamental pre-requisite towards endowing PS agents with this ability by enabling them to identify variables that describe significant features of their environment. The next step is to construct a representation and learning rule that allows the agent to identify correlations between the different variables encoded in the ECM. We conjecture that doing so will allow us to further enhance the agents' generalisation abilities. To see why, imagine that the agent is confronted with a setup $s$ such that (i) they are already strongly disposed to label $s$ with a value $v$ of some variable $V$ that is tested by an experiment $e$, and (ii) they are not strongly disposed to label $s$ with any particular value of any variable $V^{*}$ that predicts an experiment $e^{*}$. In this case, the agent will already be good at predicting the outcome of experiment $e$ when confronted with $s$, but they will not be able to reliably predict the outcome of $e^{*}$, perhaps because they haven't yet had significant experience with the $e^{*}$/$s$ pair. However, it may be that they have already noted a strong correlation between the variable $V$ and some variable $V^{*}$ which they know is predictive of $e^{*}.$ In this case, it seems that they should be able to use their knowledge regarding the value of $V$ that corresponds to $s$ to guess a corresponding value for $V^{*}$, which would then allow them to make an educated guess regarding the outcome of $e^{*}$. In future work, we aim to develop a method for identifying correlations between an agent's conceptual representations, and subsequently augment the PS learning dynamics in a way that utilises the observed correlations to allow for enhanced generalisation abilities.\footnote{This is closely related to the more general problem of identifying PS agents' subjective probability distributions over the set of identified variables. Doing this would open the door to the application of Bayesian inference methods in the context of PS, and with it, the possibility of realising Bayesian approaches to, e.g., causal inference (see, e.g., \cite{Spirtes2001Causation}), logical reasoning (see, e.g., \cite{Eva2018Bayesian}) and abduction (see, e.g., \cite{Douven1999IBE}) in PS agents.}

A second avenue for further work concerns the number of intermediate clips available to the agent. Throughout the present work, we have assumed this number to be fixed at a particular value. This is a significant assumption, which places a-priori restrictions on the kinds of abstractions that the agents are able to make. In future work, we intend to implement dynamics that allow the agent to autonomously alter its own architecture in a way that supports whatever kinds of abstraction are most useful for the learning task in which it is engaged. These dynamics would allow the agent to change the size of its label space over time as it gains information about the granularity and complexity of the hidden variables that characterise its environment. For example, one natural dynamic would be to `merge' any two labels that look like duplicates of one another (in the sense that they define very similar probability distributions over action space). Another natural dynamic would be to `split in two' any single label that is deemed to be too general and imprecise (in the sense that it defines an excessively flat probability distribution over action space). By implementing dynamics like these, we aim to make the concept formation scheme described here more autonomous, domain general and robust. 

More generally, the present work takes the first steps towards allowing PS agents to autonomously develop symbolic interfaces through which they can articulate, refine and communicate their distinctive sub-symbolic reasoning dynamics. This opens up a host of new research avenues pertaining to the further development, integration and application of such interfaces.

\subsubsection*{Acknowledgements}
This work was supported by 
(i) the Austrian Science Fund (FWF) through the SFB F71 BeyondC,
(ii) the Ministerium f{\"u}r Wissenschaft, Forschung, und Kunst Baden-W{\"u}rttemberg (AZ: 33-7533.-30-10/41/1),
(iii) the Alexander von Humboldt Foundation, and
(iv) the Zukunftskolleg of the University of Konstanz.

\bibliography{AbstractionBetaRefs} 

\end{document}